\PassOptionsToPackage{most,breakable}{tcolorbox}
\PassOptionsToPackage{dvipsnames,table}{xcolor}
\PassOptionsToPackage{numbers, compress}{natbib}
\documentclass[11pt]{article}

\usepackage[final]{ropedia}

\usepackage[T1]{fontenc}    
\usepackage[utf8]{inputenc} 
\usepackage{amsmath,amssymb}
\usepackage{booktabs}       
\usepackage{amsfonts}       
\usepackage{mathpazo}
\usepackage[xcharter,bigdelims,vvarbb]{newtxmath}
\usepackage[scaled]{helvet}
\usepackage[scaled=1.1]{zlmtt}
\usepackage{latexsym}
\usepackage{pifont}
\usepackage{placeins}
\usepackage{nicefrac}       
\usepackage{microtype}      
\usepackage{multirow}       
\usepackage{tabularx}
\usepackage{subcaption}
\usepackage{arydshln}
\usepackage{adjustbox}
\usepackage{makecell}
\usepackage{enumitem}
\usepackage{rotating}
\usepackage{titletoc}
\usepackage{comment}
\usepackage{graphicx}       
\usepackage{algorithm}      
\usepackage{algorithmic}    
\usepackage{tikz}           
\usepackage{wrapfig}
\usepackage{needspace}
\usepackage{siunitx}
\usepackage{dblfloatfix}
\usepackage{cuted}
\usepackage{capt-of}
\usepackage{xspace}
\setlist[itemize]{leftmargin=2em}

\usetikzlibrary{positioning, arrows.meta, shapes.geometric, fit, calc, backgrounds}

\definecolor{top1color}{HTML}{B19CD9} 
\definecolor{top2color}{HTML}{D4C5E8} 
\definecolor{top3color}{HTML}{E8DEEF} 
\definecolor{deltapos}{HTML}{1B7A3E}  
\definecolor{deltaneg}{HTML}{B23A48}  
\newcommand{\cone}[1]{\cellcolor{top1color}\textbf{#1}}
\newcommand{\ctwo}[1]{\cellcolor{top2color}#1}
\newcommand{\cthree}[1]{\cellcolor{top3color}#1}
\newcommand{\dpos}[1]{{\tiny\textcolor{deltapos}{\,(+#1)}}}

\firstpagelogos{}{logo/ntu_logo.jpg}{}
\newcommand{\SAgentTitleLogo}{\includegraphics[height=30pt]{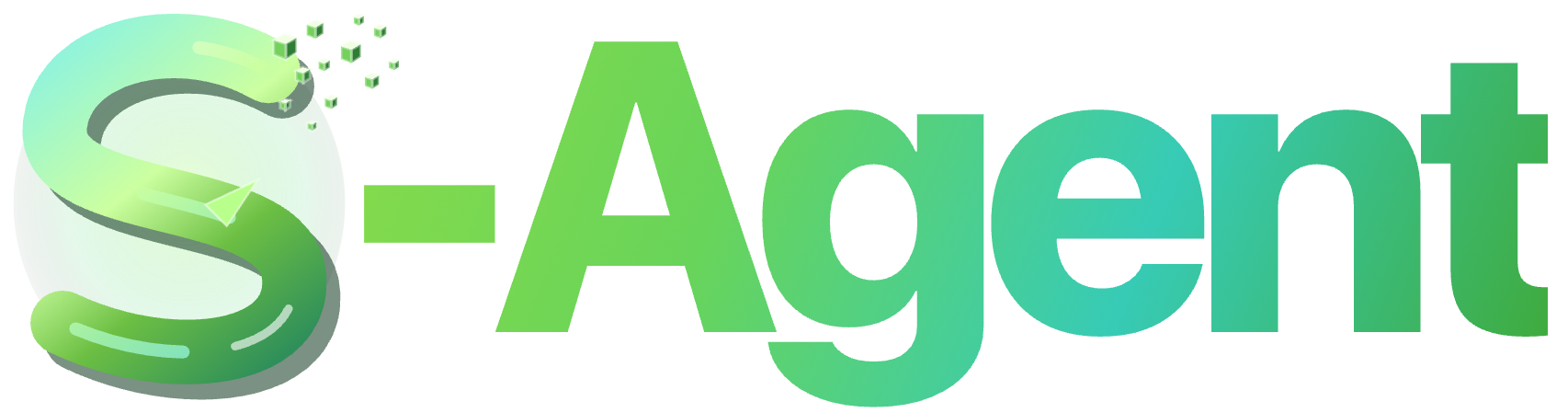}\\[-0.07in]}
\title{\SAgentTitleLogo} 
\subtitle{Spatial Tool-Use Elicits Reasoning for\\ Spatial Intelligence}

\newcommand{\coremark}{\ensuremath{^{*}}}

\newcommand{\corrmark}{\ensuremath{^{\dagger}}}

\newcommand{\authorfootnotes}{%
  \begingroup
  \renewcommand{\thefootnote}{}%
  \footnotetext{$*$ Equal contributors. $\dagger$ Corresponding author.}%
  \endgroup
  \addtocounter{footnote}{-1}%
}

\author{
  \AuthorName{Yalun Dai}{1\coremark}\quad
  \RopediaAuthorName{Hao Li}{1, \coremark}\quad
  \AuthorName{Shulin Tian}{1}\quad
  \AuthorName{Runmao Yao}{1}\quad
  \AuthorName{Yuhao Dong}{1}\\[0.55ex]
  \RopediaAuthorName{Fangzhou Hong}{1}\quad
  \RopediaAuthorName{Zhaoxi Chen}{1}\quad
  \AuthorName{Fangfu Liu}{2}\\[0.55ex]
  \AuthorName{Tao Wang}{1\corrmark}\quad
  \AuthorName{Kim-Hui Yap}{1\corrmark}\quad
  \RopediaAuthorName{Ziwei Liu}{1}\\[1.0ex]
  {\normalfont\footnotesize
  \mbox{\textsuperscript{1}NTU}\quad
  \mbox{\textsuperscript{2}THU}\quad
  \mbox{\textsuperscript{\ropediamark}Ropedia}\quad
  \paperresourceicon{\resourceprojecticon}{Project Page}{https://Ropedia.github.io/S-Agent}{Ropedia/S-Agent}}
}

\begin{document}

\maketitle
\authorfootnotes

{\centering
\includegraphics[width=0.95\textwidth]{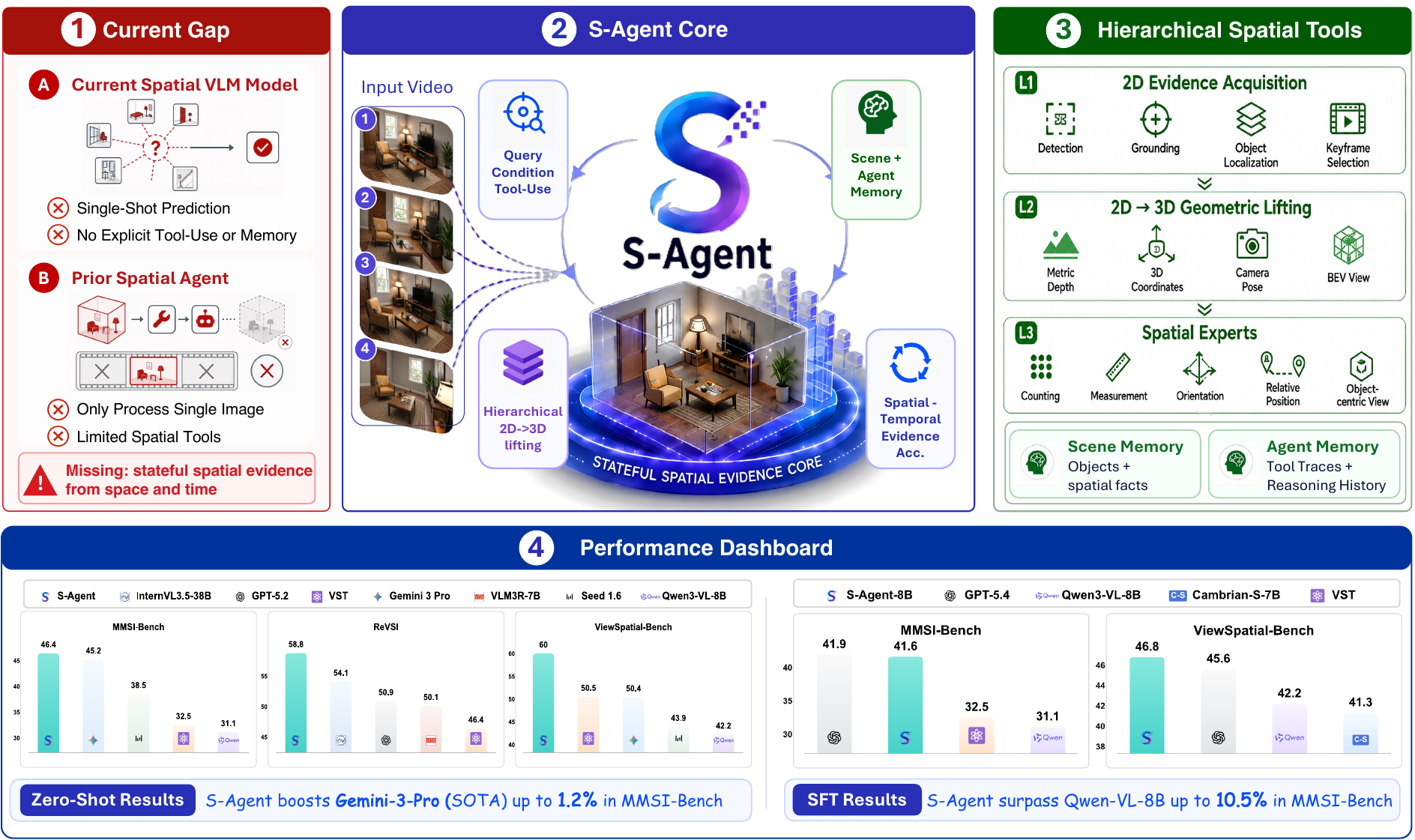}
\par
\begin{minipage}{\textwidth}
\footnotesize
\refstepcounter{figure}\label{fig:teaser}
Figure 1: \textbf{Overview of S-Agent.} S-Agent is the spatial tool-use agentic paradigm designed for continuous multi-view image and video reasoning, which formulates spatial reasoning as an active process of spatio-temporal evidence accumulation. It contains a VLM semantic planner with a hierarchy of spatial tools to ground, lift, and aggregate geometric cues, alongside a dual-memory system to maintain the evolving scene and reasoning history. Extensive experiments show that our paradigm consistently enhances zero-shot VLMs and distills a compact agent (S-Agent-8B) that rivals advanced closed-source models.
\end{minipage}
\par}

\vspace{3mm}
\begin{abstract}
Real-world spatial intelligence requires reasoning over a continuous and evolving 3D world, yet existing VLMs and tool-augmented agents largely remain tied to static, stateless inference from isolated visual observations.
We introduce \textbf{\textsc{S-Agent}}, a spatial tool-use agentic paradigm for understanding and reasoning over continuous multi-view images and videos. 
By formulating spatial reasoning as spatio-temporal evidence accumulation rather than isolated frame-level prediction, \textsc{S-Agent} reshapes spatial perception into scene-centric understanding beyond frame-centric recognition.
Specifically, \textsc{S-Agent} casts the VLM as a semantic planner that decides what evidence is needed, while a hierarchy of spatial tools and experts grounds objects in 2D, lifts them into 3D geometric evidence, and aggregates this evidence into high-level spatial knowledge (\textit{e.g.}, counting, measurement, orientation, and relative position). 
Additionally, a temporal memory mechanism, including Scene Memory for maintaining the evolving scene state and Agent Memory for accumulating reasoning context, enables evidence integration across frames and reasoning steps.
Comprehensive experiments on multi-view and video spatial reasoning benchmarks show that \textsc{S-Agent} consistently improves both open-source and closed-source VLMs in a training-free manner. 
Beyond inference-time augmentation, supervised fine-tuning (SFT) on \textsc{S-Agent}-generated spatial trajectories \textsc{S-300K} yields \textsc{S-Agent-8B}, a compact spatial agent that significantly surpasses similar-scale baselines (e.g., Qwen3-VL-8B) and performs comparably to advanced closed-source models (e.g., GPT-5.4 and Gemini 3).

\end{abstract}

\section{Introduction}
\label{sec:intro}
Spatial intelligence, the ability to understand geometric relations among objects and their 3D environments, is essential for vision-language models (VLMs) to operate in the physical world and represents a key step toward artificial general intelligence (AGI), where models are expected to perceive, reason, and make decisions in 3D space as humans do. 
Such capability is crucial for real-world applications, including embodied robotics \citep{driess2023palme,brohan2023rt2}, AR/VR perception \citep{newcombe2011kinectfusion}, and autonomous driving \citep{geiger2012kitti,caesar2020nuscenes}. 
However, unlike human perception, which naturally integrates visual cues into coherent 3D understanding, current VLMs are primarily trained on passive 2D visual-text corpora, with limited explicit 3D supervision or embodied experience \citep{radford2021clip,alayrac2022flamingo,liu2023llava}.
This creates a fundamental semantic-to-geometric gap: while VLMs excel at probabilistic and qualitative semantic inference, their reasoning is often mediated by lossy semantic representations that fail to faithfully capture high-fidelity geometry, leaving them susceptible to textual patterns and semantic priors rather than grounded 3D geometric evidence \citep{gca,testset_stress}.

Recent advances in agentic VLMs substantially push the boundary of spatial understanding by augmenting VLMs with external tools, executable programs, and explicit geometric structure. 
For example, VADAR \citep{vadar} dynamically constructs a Python API and synthesizes programs for 3D spatial reasoning; SpaceTools \citep{spacetools} trains VLMs to coordinate multiple vision and robotic tools through interactive reinforcement learning. 
However, despite their strong performance, these methods still largely focus on static images or isolated visual observations, which remains far from the goal of real-world spatial intelligence: \textit{the real 3D world is hidden, evolving, and continuously projected into streams of 2D observations}. 
Reasoning from isolated 2D views alone makes it fundamentally challenging to maintain persistent object states, integrate evidence across viewpoints and time, and build a coherent understanding of the underlying 3D scene.

To move beyond static and stateless spatial reasoning, we introduce \textbf{\textsc{S-Agent}}, a \textbf{S}patial tool-use agentic paradigm for understanding and reasoning over continuous multi-view images and videos. 
Our key motivation is that the missing ingredient for video-based spatial intelligence is not merely stronger 2D/3D visual recognition, but a reasoning mechanism that can accumulate spatial evidence along both spatial and temporal dimensions. 
Specifically, in continuous multi-view and video settings, each frame is only a partial and transient observation of the scene, while the key to spatial intelligence is to connect these observations into a spatially structured and temporally persistent understanding of the underlying 3D world.
Rather than asking VLMs to implicitly internalize this entire process, our \textsc{S-Agent} casts the VLM as a semantic planner that decides what evidence is needed, while spatial tools/experts and temporal memory provide continuous and explicit 3D awareness of the specific scene, ranging from low-level 2D/3D evidence (e.g., object grounding, depth information) to high-level spatial knowledge (e.g., orientations, relationships).
This separation enables the agent to reason from accumulated evidence instead of isolated visual impressions, extending existing spatial agent methods toward stateful, temporally grounded understanding of evolving scenes.
%

Motivated by this perspective, \textsc{S-Agent} is designed as a VLM-orchestrated spatio-temporal reasoning framework: it progressively aggregates spatial evidence from fragmented 2D observations into structured 3D scene knowledge, while persistently accumulating temporal evidence across frames and reasoning iterations. 
(1) For the \textit{spatial dimension}, \textsc{S-Agent} follows a hierarchical understanding process. 
At the first level, \textbf{2D perception tools} ground objects and regions in individual frames, establishing object-centric visual facts for subsequent reasoning. 
At the second level, \textbf{multi-view 3D tools} enrich these grounded entities with geometric cues (e.g., depth, 3D coordinates, and camera poses), allowing evidence from different viewpoints to be integrated beyond the original image plane. 
At the third level, \textbf{specialized spatial experts} aggregate these geometric signals into higher-level spatial knowledge (e.g., object counts, physical measurements, orientations, and relative positions). 
(2) For the \textit{temporal dimension}, \textsc{S-Agent} maintains memory over the evolving reasoning process: \textbf{Scene Memory} tracks grounded entities across frames to preserve object identity and suppress duplicate evidence, while \textbf{Agent Memory} stores accumulated tool observations and intermediate reasoning traces for iterative refinement. 
In this way, \textsc{S-Agent} turns video spatial reasoning from disconnected frame-level prediction into evidence accumulation over an evolving 3D scene.

Comprehensive experiments on multi-image benchmarks (MMSI-Bench~\citep{mmsi_bench} and ViewSpatial-Bench~\citep{viewspatial_bench}) and video spatial reasoning benchmarks (ReVSI~\citep{revsi} and VSI-SUPER~\citep{cambrian_s}) validate the robustness and generalizability of our approach.
(1) \textbf{Zero-shot setting}. We directly instantiate \textsc{S-Agent} with both open-source models (e.g., Qwen3) and closed-source APIs (e.g., Gemini and GPT) in a training-free manner. 
Simply and directly applying the \textsc{S-Agent} framework consistently improves the spatial reasoning ability of these VLMs, improving over GPT-5.4 by 4.5\% on MMSI-Bench. 
(2) \textbf{Training setting}. Beyond inference-time improvement, we further construct a spatial-instruction dataset \textsc{S-300K} from zero-shot \textsc{S-Agent} trajectories on the SenseNova-SI-800K \citep{sensenova_si} training set (which is fully disjoint from all evaluation benchmarks) and use it to perform supervised fine-tuning on Qwen3-VL-8B, resulting in \textsc{S-Agent-8B}. 
Compared with direct Qwen3-VL-8B inference, \textsc{S-Agent-8B} achieves a 10.5\% improvement on MMSI-Bench, improving accuracy from 31.1\% to 41.6\%, and performs comparably to advanced closed-source models such as GPT-5.4 and Gemini 3 Pro across multiple benchmarks. 
These results show that \textsc{S-Agent} is not only an effective training-free inference framework, but also a scalable paradigm for building compact spatially capable agents.

\section{Method}
\label{sec:method}


\begin{figure}[t]
    \centering
    \includegraphics[width=0.95\linewidth]{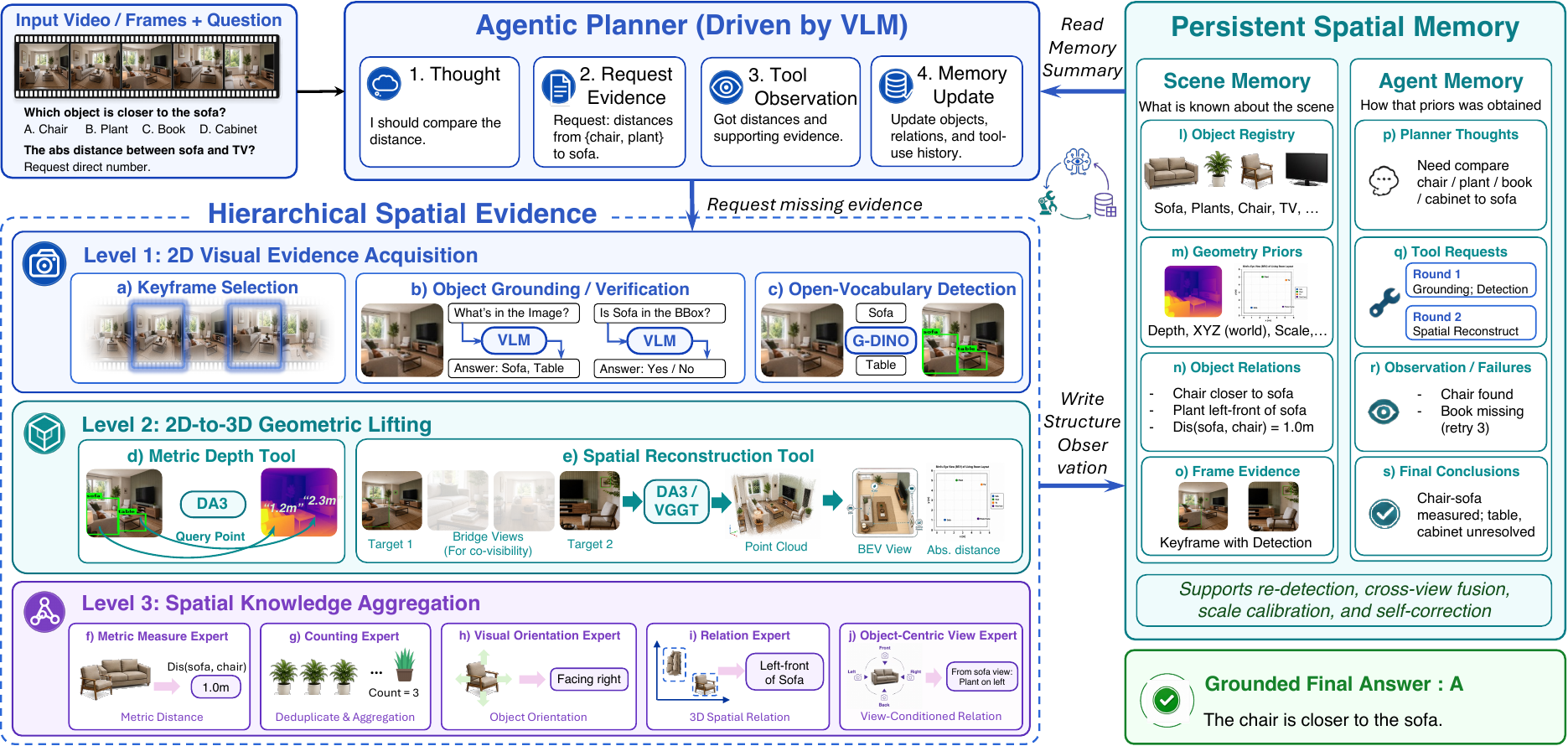}
    \caption{The pipeline of \textsc{S-Agent}. Instead of answering from an isolated visual impression, \textsc{S-Agent} uses a VLM as a semantic planner, spatial tools and experts as scene-specific evidence providers, and memory as the carrier of persistent 3D state across views, frames, and reasoning steps.}
    \label{fig:framework}
    \vspace{-2mm}
\end{figure}

%
This section details the design of \textsc{S-Agent}. 
We first formulate spatial reasoning as iterative updates to a scene state and an agent state in Section~\ref{sec:method:framework}. 
We then describe how \textsc{S-Agent} acquires hierarchical spatial evidence in Section~\ref{sec:method:spatial}, maintains temporal memory for stateful reasoning in Section~\ref{sec:method:memory}, and uses \textsc{S-Agent} trajectories to train compact agents in Section~\ref{sec:method:training}.

\subsection{S-Agent Framework}
\label{sec:method:framework}

We consider spatial reasoning problems defined by a question $q$ and a sequence or set of visual observations $\mathcal{F}$. 
The input can be a video (e.g., the scene and camera may evolve over time) or a multi-view image set (e.g., different images capture the same scene from different viewpoints). 
The goal of \textsc{S-Agent} is to produce an answer $a$ that depends on the underlying 3D scene state rather than on a single 2D projection.
To this end, \textsc{S-Agent} performs inference as an iterative evidence-seeking process, progressively acquiring and reusing scene-specific spatial evidence, as illustrated in Figure~\ref{fig:framework}.

At reasoning step $t$, \textsc{S-Agent} maintains two memory states. The first is a \emph{scene memory state} $\mathcal{S}_t$ for grounded spatial evidence, which stores grounded entities and their accumulated spatial attributes. The second is an \emph{agent memory state} $\mathcal{H}_t$ for reasoning history, which records previous tool calls, observations, and reasoning decisions. 
A tool-calling VLM planner $\pi_{\theta}$ maps the question $q$, input observations $\mathcal{F}$, and current memory states ($\mathcal{S}_t$, $\mathcal{H}_t$) to an evidence request $r_t$:
\[
    r_t = \pi_{\theta}(q, \mathcal{F}, \mathcal{S}_t, \mathcal{H}_t).
\]
A spatial tool or expert executes $r_t$ and returns an observation $o_t$, which is used to update both memory states:
\[
    (\mathcal{S}_{t+1}, \mathcal{H}_{t+1}) =
    \mathrm{Update}(\mathcal{S}_t, \mathcal{H}_t, r_t, o_t).
\]
The agent terminates when the accumulated evidence is sufficient to answer $q$. This formulation separates \emph{semantic planning} from \emph{spatial evidence acquisition}: the VLM decides what to measure or compare, while tools and memory provide scene-specific spatio-temporal evidence for the final reasoning.
Unlike fixed pipelines or standard tool-use agents that treat each tool call as an isolated action, \textsc{S-Agent} conditions each evidence request on both the question and the evolving memory state. 
As a result, perception and geometric computation are invoked on demand, and their outputs remain reusable across later reasoning steps. 
The following sections describe the two core mechanisms of this framework: hierarchical spatial evidence acquisition (Section~\ref{sec:method:spatial}) and temporal memory for stateful reasoning (Section~\ref{sec:method:memory}).

\subsubsection{Hierarchical Spatial Evidence}
\label{sec:method:spatial}

\textsc{S-Agent} acquires spatial evidence through a three-level hierarchy that transforms raw 2D observations into explicit, scene-specific spatial knowledge. 
This hierarchy reflects the varying levels of evidence required by spatial tasks: some questions can be answered from localized image-level cues, while others require lifting those cues into 3D geometry or aggregating them through specialized spatial experts. 
This staged design keeps the VLM focused on semantic planning, while delegating scene-specific perception and spatial computation (e.g., visual localization, geometric recovery, and metric or relational computation) to tools whose outputs can also be stored and reused in memory.

We denote the three tool levels as $\mathcal{T}^{(1)}$, $\mathcal{T}^{(2)}$, and $\mathcal{T}^{(3)}$, corresponding to 2D visual evidence acquisition, 2D-to-3D geometric lifting, and spatial knowledge aggregation, respectively. 
Given an evidence request $r_t$, \textsc{S-Agent} selects a tool or expert $T^{(k)} \in \mathcal{T}^{(k)}$ and produces an observation
\[
    o_t = T^{(k)}(r_t, \mathcal{F}, \mathcal{S}_t), \quad k \in \{1,2,3\}.
\]
Depending on the selected level, $o_t$ may contain localized image-level cues, lifted 3D geometry, or high-level spatial knowledge.

\textbf{Level 1: 2D Visual Evidence Acquisition} (Figure~\ref{fig:framework}(a-c)).
The first level identifies what visual evidence should be extracted from the raw 2D observations before higher-level spatial reasoning. 
Since videos or multi-view images contain many redundant, partial, or irrelevant views, \textsc{S-Agent} first gathers query-relevant image-level cues, such as selecting informative frames, grounding referred entities with VLMs, and localizing candidate regions with open-vocabulary detectors. 
These image-level cues can directly support simple queries, while also serving as observations for subsequent 3D lifting and spatial reasoning.

\textbf{Level 2: 2D-to-3D Geometric Lifting} (Figure~\ref{fig:framework}(d-e)). 
The second level lifts image-level evidence into a 3D-aware representation of the scene. 
Given the cues collected at Level 1, \textsc{S-Agent} invokes multi-view geometric tools to recover scene-level 3D information, such as depth structure, metric coordinates, camera poses, and bird's-eye-view or novel-view evidence.
This geometric lifting allows the agent to reason beyond the original image plane: fragmented 2D observations can be compared in a shared spatial context, apparent 2D size can be disambiguated from physical scale, and spatial relations can be evaluated with respect to camera motion or alternative viewpoints.

\textbf{Level 3: Spatial Knowledge Aggregation} (Figure~\ref{fig:framework}(f-j)).
The third level abstracts the 2D and 3D cues collected in the previous stages into high-level, scene-specific spatial knowledge. 
To this end, \textsc{S-Agent} uses a set of specialized spatial experts, each responsible for a particular class of spatial queries, including counting, relative direction, object orientation, and physical size/distance. 
These experts aggregate the relevant evidence and return structured observations that can be directly consumed by the VLM planner for final reasoning. 
This design turns fragmented perceptual and geometric cues into explicit scene-level spatial knowledge, reducing the need for the VLM to perform unreliable metric or relational reasoning in free-form text.

Details of the tools and experts used in Levels 1-3 are provided in Appendix~\ref{app:tools}.

\subsubsection{Temporal Memory for Stateful Reasoning}
\label{sec:method:memory}
To support stateful reasoning over continuous observations, \textsc{S-Agent} maintains two complementary memories: \textit{Scene Memory} for reusable scene evidence and \textit{Agent Memory} for the reasoning process. 
Each tool or expert observation from Section~\ref{sec:method:spatial} updates both memories in different ways: its scene-relevant content is consolidated into Scene Memory, while the request, returned observation, and reasoning context are recorded in Agent Memory. 
This separation allows the VLM planner to reason over accumulated spatial knowledge while keeping track of what has been tried, what remains uncertain, and what evidence should be requested next.

Formally, after executing request $r_t$ and receiving observation $o_t$, each tool observation is decomposed into reusable scene evidence $e_t$ and process context $c_t$. 
The two memories are then updated with different operations:
\vspace{-2mm}
\[
    \mathcal{S}_{t+1} = \operatorname{Merge}(\mathcal{S}_t, e_t),
    \qquad
    \mathcal{H}_{t+1} = \operatorname{Append}(\mathcal{H}_t, c_t).
\]
Scene Memory merges $e_t$ into the current scene state, either by updating an existing entry or creating a new one, while Agent Memory appends $c_t$ to the reasoning trajectory.

\textbf{Scene Memory} (Figure~\ref{fig:framework}(l-o)).
Scene Memory turns 2D/3D cues into a persistent, scene-level understanding. 
In multi-view images or videos, the same object may appear across different frames, viewpoints, scales, and referring expressions. Without a persistent memory, reasoning over these cues independently would lead to duplicated evidence and unstable object identity.
Scene Memory therefore consolidates scene-relevant tool/expert observations into an evolving, entity-centric memory, binding repeated observations to persistent scene entities and accumulating their visual and geometric evidence over time. 
It is not a dense reconstruction of the full environment, but a question-conditioned spatial memory that preserves the evidence needed for the current query.

Concretely, Scene Memory stores two types of reusable content: grounded entities and derived spatial facts.
For entities, the memory stores their textual aliases, supporting frames, localized visual evidence, and accumulated geometric attributes. 
For derived facts, it stores spatial relations or measurements computed by higher-level experts, together with the evidence from which they are derived.
When a new observation arrives, \textsc{S-Agent} either links it to an existing scene memory entry or creates a new one, allowing later reasoning steps to reuse previously grounded evidence or facts instead of re-processing each frame from scratch.

\textbf{Agent Memory} (Figure~\ref{fig:framework}(p-s)).
Agent Memory preserves the reasoning process that leads to the evolving scene understanding. 
In iterative tool-use reasoning, the agent should remember not only what has been observed, but also what has already been tried, which evidence was requested, which tools succeeded or failed, and why the planner decided to continue.
Without such process memory, the planner may repeatedly issue redundant tool calls, overlook unresolved uncertainties, or contradict its earlier observations.
Agent Memory therefore records the reasoning trajectory across iterations, providing the planner with a compact context for deciding the next evidence request.

Specifically, Agent Memory stores the planner's intermediate thoughts, issued tool calls, returned observations, failure messages, and intermediate conclusions.
Unlike Scene Memory, which consolidates reusable scene evidence, Agent Memory keeps the procedural context around how that evidence was obtained and used. 
When the planner receives a new memory summary, it can identify missing evidence, revisit uncertain observations, or refine its strategy based on previous tool feedback.


\subsection{Training-Time Distillation}
\label{sec:method:training}
Beyond inference-time reasoning, \textsc{S-Agent} can also serve as a teacher for training compact spatial agents. 
We construct training data from SenseNova-SI-800K \citep{sensenova_si} by selecting samples that are both challenging for a weaker student model and likely to require tool use. 

\textbf{Data generation.}
We estimate sample difficulty from multiple rollouts of Qwen3-VL-8B and prioritize questions on which the student is uncertain or unstable, rather than questions it already solves reliably. 
We further favor spatial questions that are likely to benefit from tool use, such as metric measurement, counting, relative position, camera/viewpoint reasoning, and grounding-dependent queries. 
A frozen teacher \textsc{S-Agent}, instantiated with GPT-5.4, is then used to generate complete trajectories, including planner prompts and responses, tool calls, tool observations, intermediate artifacts, memory states, final answers, and evaluation results.

\begin{wrapfigure}{r}{0.75\textwidth}
    \centering
    \vspace{-0.8em}
    \includegraphics[width=0.73\textwidth]{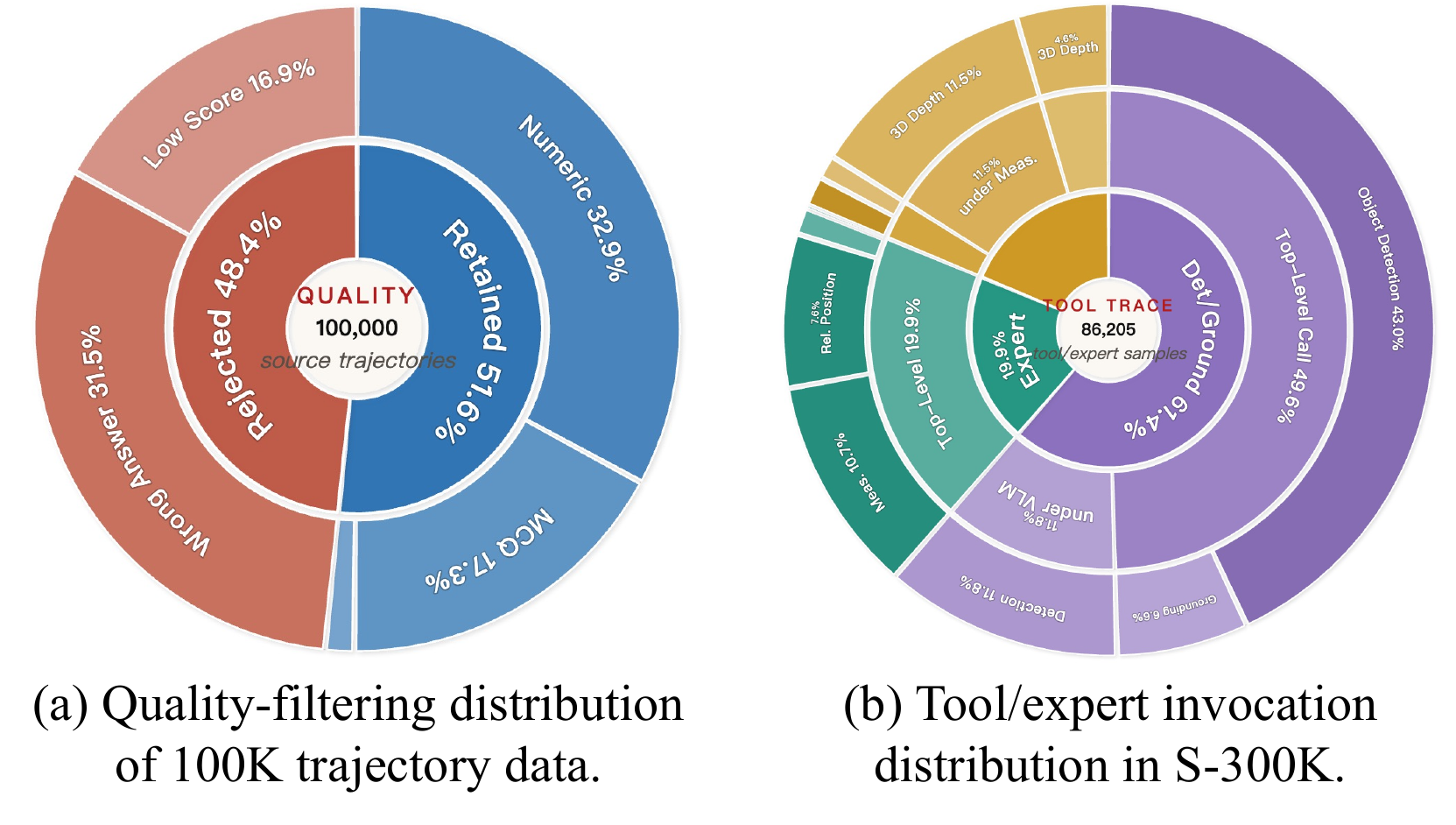}
    \vspace{-0.2in}
    \caption{Data composition and tool invocation statistics of \textbf{S-300K}.}
    \vspace{-0.1in}
    \label{fig:s300k}
\end{wrapfigure}

\textbf{Data filtering.}
We then apply quality filtering when exporting trajectories for supervised fine-tuning. 
All generated trajectories are first preserved in full as raw agent traces for analysis and possible re-export, regardless of whether the final answer is correct or whether some tool calls fail. 
For SFT data, we retain only trajectories with valid executions and correct final answers under answer-type-specific criteria. 
Multiple-choice questions require the predicted option in the final answer to match the ground-truth option, numeric questions are filtered by mean relative accuracy, and text questions are filtered by normalized answer matching. 
Importantly, tool usage itself is not used as a hard filtering criterion: the goal is to keep high-quality agent behavior while allowing the planner to decide when tool calls are necessary.
The filtering ratio distribution is shown in Figure~\ref{fig:s300k}(a).

\textbf{Data decomposition.}
Each retained trajectory is finally decomposed into multiple forms of supervision rather than being used only as a final-answer example.
We construct final-answer trajectories to teach end-to-end spatial reasoning, turn-level trajectories to teach iterative tool-use decisions under partial reasoning context, and expert/tool trajectories to improve spatial tool-use policy and expert-level reasoning.
This decomposition converts a single teacher-agent rollout into multi-granularity training signals, enabling the student model to learn not only the final answer distribution, but also how to request evidence, interpret tool observations, and accumulate spatial knowledge across reasoning steps. 

After this process, we obtain the \textsc{S-300K} dataset for supervised fine-tuning. 
We fine-tune Qwen3-VL-8B on \textsc{S-300K} to obtain our compact spatial agent, \textsc{S-Agent-8B}. 
The detailed data distribution of \textsc{S-300K} is shown in Figure~\ref{fig:s300k}(b).
Further details are provided in Appendix~\ref{app:training_data}.





\section{Experiments}
\label{sec:experiments}
We conduct extensive experiments on a diverse suite of spatial reasoning benchmarks to evaluate \textsc{S-Agent} under both training-free zero-shot and trained-agent regimes. 
Section~\ref{sec:experiments:setup} introduces the training and evaluation setup. 
Section~\ref{sec:experiments:zeroshot} reports the main zero-shot and comparative results. 
Section~\ref{sec:experiments:training} evaluates training compact agents from \textsc{S-Agent} trajectories. 
Section~\ref{sec:experiments:ablation} presents ablations, and Section~\ref{sec:experiments:analysis} analyzes qualitative examples and failure cases.

\begin{table*}[t]
    \centering
    \small
    \setlength{\tabcolsep}{3pt}
    \caption{
    Detailed MMSI-Bench results. We follow the taxonomy of~\citep{mmsi_bench} and group dimensions into Positional Relationship, Geometric Attribute, Motion Perception, and Multi-step Reasoning (MSR). C/O/R denote camera/object/region in positional-relation subcategories. \textit{SenseNova} is abbreviated as \textit{SN}. Top-1/2/3 results are highlighted in \colorbox{top1color}{\textbf{deep}}, \colorbox{top2color}{medium}, and \colorbox{top3color}{light} lavender.}
    \resizebox{\textwidth}{!}{%
    \begin{tabular}{lcccccc|cc|cc|c|c}
        \toprule
        \multirow{2}{*}{\textbf{Model}} & \multicolumn{6}{c}{\textbf{Positional Relationship}} & \multicolumn{2}{c}{\textbf{Geometric Attr.}} & \multicolumn{2}{c}{\textbf{Motion Perception}} & \multirow{2}{*}{\textbf{MSR}} & \multirow{2}{*}{\textbf{Avg.}} \\
        \cmidrule(lr){2-7} \cmidrule(lr){8-9} \cmidrule(lr){10-11}
        & \textbf{C-C} & \textbf{O-O} & \textbf{R-R} & \textbf{C-O} & \textbf{O-R} & \textbf{C-R} & \textbf{Meas.} & \textbf{Appr.} & \textbf{Cam.} & \textbf{Obj.} & & \\
        \midrule
        \multicolumn{13}{l}{\textit{Proprietary Models}} \\
        Gemini 3 Pro & \ctwo{47.3} & \cone{48.9} & \cone{42.0} & 43.0 & 37.6 & \cthree{60.2} & \cone{64.1} & \cthree{39.4} & \ctwo{41.9} & \ctwo{47.4} & 37.9 & \ctwo{45.2} \\
        Gemini 2.5 Pro & 38.7 & 34.0 & \ctwo{40.7} & 44.2 & 38.8 & 41.0 & \ctwo{62.5} & 30.3 & 39.2 & 25.0 & 33.3 & 38.0 \\
        GPT-5.4 & 41.9 & 33.0 & 35.8 & 49.8 & \cthree{42.4} & \cone{68.7} & 54.7 & 37.4 & 28.3 & 40.8 & 36.4 & \cthree{41.9} \\
        Grok 4 & 36.6 & 35.1 & \cthree{39.5} & 34.9 & \cone{45.9} & 50.6 & 21.9 & 22.7 & \cthree{40.5} & \cthree{43.4} & \cthree{38.4} & 37.8 \\
        \midrule
        \multicolumn{13}{l}{\textit{Open-weight General Models}} \\
        Seed 1.6 & 36.6 & 36.2 & 32.1 & 32.6 & \cthree{42.4} & 46.9 & 48.4 & 33.0 & 31.1 & 42.1 & \ctwo{40.4} & 38.5 \\
        InternVL3\_5-8B             & 29.0 & 26.6 & 29.6 & 24.4 & 31.8 & 25.3 & 29.7 & 25.8 & 14.9 & 34.2 & 36.4 & 29.0 \\
        SN-U1-8B-MoT & \cthree{46.2} & \cthree{41.5} & 29.6 & \ctwo{58.1} & 38.8 & \ctwo{63.9} & 43.8 & 21.2 & 25.7 & 31.6 & 26.8 & 38.0 \\
        Qwen3-VL-8B-Instruct & 28.0 & 37.2 & 32.1 & 31.4 & 35.3 & 38.5 & 37.5 & 15.2 & 27.0 & 28.9 & 29.8 & 31.1 \\
        Qwen3-VL-8B-Thinking & 31.2 & 26.6 & 32.1 & 29.1 & 32.9 & 30.1 & 50.0 & 16.7 & 17.6 & 23.7 & 27.3 & 28.6 \\
        Qwen3.5-9B & 34.4 & 36.2 & 34.6 & 39.5 & 38.8 & 54.2 & 56.3 & 28.8 & 36.5 & 26.3 & 28.8 & 36.5 \\
        Qwen3-VL-30B-A3B-Thinking & 23.7 & 31.9 & 35.8 & 31.4 & 36.5 & 22.9 & 40.6 & 19.7 & 18.9 & 27.6 & 31.3 & 29.4 \\
        \midrule
        \multicolumn{13}{l}{\textit{Open-weight Spatial Models}} \\
        SN-SI-1.1-Qwen2.5VL-7B & \cone{51.6} & 29.8 & 32.1 & \cthree{50.0} & 29.4 & 42.2 & 37.5 & 28.8 & 23.0 & 34.2 & 18.7 & 32.8 \\
        SN-SI-1.1-Qwen3VL-8B & 44.1 & 38.3 & 33.3 & \cone{65.1} & 38.8 & 59.0 & 48.4 & 24.2 & 29.7 & 34.2 & 22.2 & 38.1 \\
        VST-7B-SFT & 39.8 & 36.2 & 35.8 & 37.2 & 29.4 & 33.7 & 29.7 & \cone{47.0} & 36.5 & 35.5 & 18.2 & 32.5 \\
        \midrule
        \textbf{Ours (S-Agent)} & \cthree{46.2}\dpos{17.2} & \ctwo{43.6}\dpos{17.0} & 37.0\dpos{7.4} & 43.0\dpos{18.6} & \ctwo{43.5}\dpos{11.7} & \ctwo{63.9}\dpos{38.6} & \cthree{57.8}\dpos{28.1} & \ctwo{40.9}\dpos{15.1} & \cone{46.0}\dpos{31.1} & \cone{48.7}\dpos{14.5} & \cone{44.4}\dpos{8.0} & \cone{46.4}\dpos{17.4} \\
        \bottomrule
    \end{tabular}%
    }
    \vspace{2mm}
    \label{tab:mmsi_bench_zeroshot}
\end{table*}

\subsection{Experimental Setup}
\label{sec:experiments:setup}
\textbf{Benchmarks.}
We evaluate \textsc{S-Agent} on four benchmarks that stress different forms of spatial reasoning across multi-image and video inputs. 
For \textit{multi-image} reasoning, \textbf{MMSI-Bench}~\citep{mmsi_bench} provides multiple images of the same scene and tests whether models can integrate evidence across views for positional relationships, geometric attributes, motion perception, and multi-step spatial reasoning. 
\textbf{ViewSpatial-Bench}~\citep{viewspatial_bench} focuses more specifically on perspective-aware localization, requiring models to localize objects or infer positions under different camera viewpoints. 
For \textit{video reasoning}, \textbf{ReVSI}~\citep{revsi} evaluates 3D spatial reasoning from dynamic observations, emphasizing whether models can infer spatial relations that are not reliably recoverable from isolated frames. 
\textbf{VSI-SUPER}~\citep{cambrian_s} focuses on video spatial change reasoning, requiring models to identify how objects, viewpoints, or spatial layouts change over time.

\textbf{Baselines.}
We compare \textsc{S-Agent} with three categories of baselines: advanced proprietary VLMs (e.g., \textit{Gemini 3 Pro, GPT-5.4, and Grok 4}), open-weight general VLMs (e.g., \textit{Qwen series}), and spatially specialized models (e.g., \textit{Cambrian-S, VST-SFT, and SenseNova-SI series}). 
The first two groups measure performance against strong general-purpose multimodal systems, while the third evaluates whether \textsc{S-Agent} can compete with models explicitly trained or tuned for spatial reasoning.

\textbf{Models.} 
In the zero-shot setting, we instantiate \textsc{S-Agent} with advanced VLMs (GPT-5.4 and Gemini 3 Pro) as tool-calling planners, without any task-specific training. 
In the trained-agent setting, we use Qwen3-VL-8B-Instruct as the backbone planner and train it on trajectories generated by zero-shot \textsc{S-Agent}, yielding our compact agent \textsc{S-Agent-8B}.

\textbf{Training Data.}
We construct training data from SenseNova-SI-800K~\citep{sensenova_si}, which is fully disjoint from all evaluation benchmarks. 
We randomly sample 100K questions and use zero-shot \textsc{S-Agent} with a GPT-5.4 planner to generate tool-use trajectories.
We then filter trajectories by execution validity and final-answer correctness, and decompose the retained trajectories into full final-answer samples, turn-level VLM-call samples, and expert/tool-specific samples. 
This yields 292,391 SFT samples, denoted as \textsc{S-300K}. 
Appendix~\ref{app:training_data} provides the detailed filtering criteria and data distribution.

\textbf{Training Configuration.}
We fine-tune Qwen3-VL-8B-Instruct on \textsc{S-300K} using LLaMA-Factory \citep{zheng2024llamafactory} with the \texttt{qwen3\_vl\_nothink} template on $8\times$ B200 GPUs. 
The model is trained with the standard supervised next-token prediction objective over assistant responses, including serialized tool-use trajectories, tool observations, and final answers. 
We use a maximum sequence length of 8192, a learning rate of $5\times10^{-5}$, cosine learning-rate decay with 3\% warmup, and train for one epoch. 
The resulting compact spatial agent is denoted as \textsc{S-Agent-8B}.

\begin{table*}[t]
    \centering
    \scriptsize
    \setlength{\tabcolsep}{3pt}
    \caption{Results on ViewSpatial-Bench~\cite{viewspatial_bench}. We report the official five question types: camera-perspective object view orientation (C-OVO), camera-perspective relative direction (C-RD), person-perspective object view orientation (P-OVO), person-perspective relative direction (P-RD), and person-perspective scene-simulation relative direction (P-SSRD).} 
    \label{tab:viewspatial_zeroshot}
    \begin{tabular}{lcc|ccc|c}
        \toprule
        \multirow{2}{*}{\textbf{Model}} & \multicolumn{2}{c}{\textbf{Camera Perspective}} & \multicolumn{3}{c}{\textbf{Person Perspective}} & \multirow{2}{*}{\textbf{Avg.}} \\
        \cmidrule(lr){2-3} \cmidrule(lr){4-6}
        & \textbf{C-OVO} & \textbf{C-RD} & \textbf{P-OVO} & \textbf{P-RD} & \textbf{P-SSRD} & \\
        \midrule
        \multicolumn{7}{l}{\textit{Proprietary Models}} \\
        Gemini 3 Pro & 31.6 & \ctwo{61.9} & 41.1 & \ctwo{74.4} & 38.9 & 50.4 \\
        Gemini 2.5 Pro & 33.0 & 59.1 & 51.0 & 45.8 & 32.6 & 46.1 \\
        GPT-5.4 & 27.9 & 60.2 & 41.0 & 48.5 & 40.1 & 45.6 \\
        Grok-4 & 23.9 & 57.1 & 47.6 & 51.7 & 24.9 & 43.2 \\
        \midrule
        \multicolumn{7}{l}{\textit{Open-weight General Models}} \\
        Seed 1.6 & 26.9 & 55.8 & 54.8 & 48.5 & 26.6 & 43.9 \\
        Qwen3-VL-8B-Instruct & 29.7 & 54.2 & 47.3 & 40.3 & 31.1 & 42.2 \\
        BAGEL-7B-MoT & \ctwo{38.7} & 48.3 & 47.0 & 42.5 & 26.5 & 41.3 \\
        InternVL3\_5-8B & 24.7 & 49.8 & 50.3 & 34.6 & 32.9 & 40.0 \\
        \midrule
        \multicolumn{7}{l}{\textit{Open-weight Spatial Models}} \\
        Cambrian-S-7B & 22.7 & 50.4 & 45.0 & 38.8 & 41.9 & 41.3 \\
        VST-3B-SFT & \cthree{35.4} & 46.9 & \cone{70.3} & \cthree{52.6} & \ctwo{62.8} & \ctwo{52.9} \\
        VST-7B-SFT & 29.6 & 52.7 & 51.9 & 50.7 & \cone{64.5} & 50.5 \\
        SN-SI-1.1-Qwen2.5VL-7B & 26.7 & 47.9 & \cthree{57.1} & 43.2 & 49.7 & 45.5 \\
        SN-SI-1.1-Qwen3VL-8B & 22.0 & \cthree{60.3} & \ctwo{67.8} & 41.5 & 55.6 & \cthree{51.2} \\
        \midrule
        \textbf{Ours (\textsc{S-Agent})} & \cone{55.5}\dpos{27.6} & \cone{62.5}\dpos{2.3} & 42.2\dpos{1.2} & \cone{81.1}\dpos{32.6} & \cthree{60.6}\dpos{20.5} & \cone{60.0}\dpos{14.4} \\
        \bottomrule
    \end{tabular}%
    \vspace{-2mm}
\end{table*}

\vspace{-2mm}
\subsection{Zero-Shot Performance.}
\label{sec:experiments:zeroshot}
We report the results on MMSI-Bench, ViewSpatial-Bench, and ReVSI in the main text, while the results on VSI-SUPER are provided in Appendix~\ref{app:more_experiments}.

\textbf{Results on MMSI-Bench.}
Table~\ref{tab:mmsi_bench_zeroshot} shows that our \textsc{S-Agent} achieves the best overall zero-shot performance on MMSI-Bench, obtaining the highest average score of 46.4\%. 
It outperforms the strongest proprietary baseline Gemini 3 Pro by 1.2\%, and surpasses GPT-5.4 by 4.5\%.
Notably, \textsc{S-Agent} achieves the best results on both motion perception subtasks, i.e., camera motion (46.0\%) and object motion (48.7\%), as well as multi-step reasoning (44.4\%), while remaining competitive across positional and geometric categories.
These results demonstrate the effectiveness of \textsc{S-Agent} for zero-shot spatial reasoning, with particularly strong performance on dynamic motion understanding and multi-step reasoning while maintaining robust results across static spatial and geometric tasks.

\textbf{Results on ViewSpatial-Bench.}
Table~\ref{tab:viewspatial_zeroshot} reports the zero-shot results on ViewSpatial-Bench. \textsc{S-Agent} achieves an average score of 60.0\%, outperforming GPT-5.4 by 14.4\%. 
It obtains the best performance on C-OVO (55.5\%) and P-RD (81.1\%), showing strong capability in both camera-centered and person-centered spatial reasoning. 
\textsc{S-Agent} also brings large gains on the more challenging P-SSRD split, improving over GPT-5.4 by 20.5\%. These results further demonstrate the effectiveness of \textsc{S-Agent} for zero-shot view-aware spatial reasoning, especially when reasoning over relative directions and perspective-dependent spatial relations.

\textbf{Results on ReVSI.}
Table~\ref{tab:revsi_zeroshot} reports detailed results on ReVSI.
\textsc{S-Agent} achieves an average score of 58.8, ranking second overall and outperforming all open-source general models and spatially specialized baselines.
The gains are especially pronounced on multiple-choice spatial reasoning tasks: \textsc{S-Agent} obtains the best results on relative direction and route planning, and ranks third on relative distance.
These categories require integrating evidence across frames and viewpoints rather than relying on a single visual impression, which aligns well with the design of stateful evidence accumulation.

\begin{table*}[t]
    \centering
    \small
    \setlength{\tabcolsep}{3pt}
    \providecommand{\vsiscore}[1]{{\tiny\textcolor{gray}{\, (#1)}}}
    \caption{Detailed comparison on the ReVSI~\cite{revsi} leaderboard. ReVSI scores are shown as the main values, and corresponding VSI-Bench scores from the official ReVSI experiments page are shown in gray parentheses when available. We follow the official evaluation dimensions: four numerical question types (object counting, absolute distance, object size, and room size) and three multiple-choice question types (relative distance, relative direction, and route planning). The top-1 / top-2 / top-3 ReVSI results in each column, excluding chance baselines, are highlighted with \colorbox{top1color}{\textbf{deep}}, \colorbox{top2color}{medium}, and \colorbox{top3color}{light} lavender.}
    \vspace{2mm}
    \label{tab:revsi_zeroshot}
    \resizebox{\textwidth}{!}{%
    \begin{tabular}{lc|cccc|ccc|c}
        \toprule
        \multirow{2}{*}{\textbf{Model}} & \multirow{2}{*}{\textbf{Frames}} & \multicolumn{4}{c}{\textbf{Numerical Question}} & \multicolumn{3}{c}{\textbf{Multiple-Choice Question}} & \multirow{2}{*}{\textbf{Avg.}} \\
        \cmidrule(lr){3-6} \cmidrule(lr){7-9}
        & & \textbf{Obj. Cnt.} & \textbf{Abs. Dist.} & \textbf{Obj. Size} & \textbf{Room Size} & \textbf{Rel. Dist.} & \textbf{Rel. Dir.} & \textbf{Route Plan} & \\
        \midrule
        \multicolumn{10}{l}{\textit{Baseline}} \\
        Chance (Random) & ALL & -- & -- & -- & -- & 23.7 & 26.8 & 26.0 & -- \\
        Chance (Frequency) & ALL & 52.2 & 40.1 & 17.4 & 20.9 & 25.8 & 31.9 & 30.2 & 31.4 \\
        \midrule
        \multicolumn{10}{l}{\textit{Proprietary Models (API)}} \\
        GPT-5.2 & 64 & \cthree{56.2} & 41.5 & \cthree{73.9} & \cone{63.0} & 48.4 & 34.9 & 38.2 & 50.9 \\
        Gemini 3 Flash & 1 FPS & \cone{65.7} & 53.1 & \ctwo{77.6} & 52.8 & \ctwo{64.6} & 47.9 & 41.8 & \cthree{57.6} \\
        Gemini 3 Pro & 1 FPS & \ctwo{60.1} & 54.7 & \cone{79.3} & 51.9 & \cone{68.1} & \ctwo{56.0} & \ctwo{56.4} & \cone{60.9} \\
        \midrule
        \multicolumn{10}{l}{\textit{Open-Source General Models}} \\
        Qwen3-VL-8B-Instruct & 64 & 40.4 & 52.3 & 69.0 & 45.1 & 57.1 & 39.5 & 40.5 & 49.1 \\
        Qwen3-VL-32B-Instruct & 64 & 46.9 & \cone{65.0} & 70.4 & \cthree{55.8} & 53.8 & 34.0 & \cthree{47.3} & 53.3 \\
        InternVL3.5-8B & 64 & 43.3 & 54.6 & 64.2 & 47.6 & 45.0 & 36.3 & 44.4 & 47.9 \\
        InternVL3.5-38B & 64 & 43.8 & \cthree{60.6} & 70.2 & \ctwo{58.4} & 57.4 & 45.9 & 42.7 & 54.1 \\
        LLaVA-Video-7B-Qwen2 & 64 & 31.3 & 1.4 & 52.5 & 16.7 & 38.3 & 33.3 & 38.4 & 30.3 \\
        LLaVA-Video-72B-Qwen2 & 64 & 40.1 & 29.6 & 59.3 & 27.9 & 39.6 & 24.8 & 43.0 & 37.8 \\
        \midrule
        \multicolumn{10}{l}{\textit{Spatially Specialized Models and Base Models}} \\
        Cambrian-S-7B & 128 & 48.4 & 60.5 & 65.5 & 46.7 & 37.1 & 48.5 & 37.0 & 49.1 \\
        Qwen2.5-VL-7B-Instruct & 4 FPS & 36.9 & 15.0 & 49.7 & 29.0 & 31.5 & 29.5 & 36.7 & 32.6 \\
        VST-7B-SFT & 4 FPS & 35.4 & 52.6 & 67.9 & 47.2 & 49.2 & 36.9 & 35.4 & 46.4 \\
        Qwen2.5-VL-7B-Instruct & 32 & 34.3 & 21.7 & 45.5 & 35.1 & 32.6 & 33.7 & 34.1 & 33.9 \\
        SpaceR-7B (SG-RLVR) & 32 & 30.7 & 34.5 & 52.0 & 18.6 & 22.8 & 34.5 & 20.2 & 30.5 \\
        Qwen2.5-VL-3B-Instruct & 16 & 18.7 & 15.6 & 16.8 & -- & 33.2 & 34.3 & -- & 23.7 \\
        Spatial-MLLM-4B-135k & 16 & 40.7 & 45.3 & 46.8 & -- & 32.3 & 37.4 & -- & 40.5 \\
        Spatial-MLLM-4B-820k & 16 & 41.5 & 40.0 & 53.1 & -- & 30.7 & 39.2 & -- & 40.9 \\
        LLaVA-Video-7B-Qwen2 & 32 & 29.9 & 1.5 & 53.0 & 19.3 & 39.1 & 33.8 & 38.8 & 30.8 \\
        VLM3R-7B & 32 & 41.6 & \ctwo{61.6} & 64.8 & 52.5 & 46.5 & \cthree{49.5} & 34.1 & 50.1 \\
        \midrule
        \textbf{Ours (S-Agent)} & 64 & 54.0 & 45.6 & 62.6 & 53.4 & \cthree{63.6} & \cone{66.4} & \cone{66.1} & \ctwo{58.8} \\
        \bottomrule
    \end{tabular}%
    }
\end{table*}

\vspace{-2mm}
\begin{table*}[t]
\centering
\small
\begin{minipage}[t]{0.50\textwidth}
\centering
\setlength{\tabcolsep}{3pt}
\caption{Trajectory distillation results across three main spatial reasoning benchmarks.} 
\vspace{0.15cm}
\label{tab:trajectory_distillation}
\resizebox{\linewidth}{!}{%
\begin{tabular}{lccc}
    \toprule
    \textbf{Model} & \textbf{MMSI} & \textbf{ViewSpatial} & \textbf{ReVSI} \\
    \midrule
    \multicolumn{4}{l}{\textit{Proprietary VLMs}} \\
    Gemini 3 Pro & 45.2 & 50.4 & 60.9  \\
    GPT-5.4 & 41.9 & 45.6 & -  \\
    \midrule
    \multicolumn{4}{l}{\textit{Open-weight Models}} \\
    Qwen3-VL-8B-Instruct & 31.1 & 42.2 & 49.1  \\
    \textsc{S-Agent} (Qwen3-VL-8B) & 30.7 & 44.1 & 49.5   \\
    \textbf{\textsc{S-Agent-8B}} & \textbf{41.6} & \textbf{46.8} & \textbf{52.8}  \\
    \bottomrule
\end{tabular}%
}
\end{minipage}
\hfill
\begin{minipage}[t]{0.45\textwidth}
\centering
\setlength{\tabcolsep}{4pt}
\caption{Ablation on ViewSpatial with \textsc{S-Agent} using GPT-5.4 as the planner.}
\label{tab:ablation_config}
\resizebox{\linewidth}{!}{%
\begin{tabular}{lccccc|c}
\toprule
\multirow{2}{*}{\textsc{S-Agent}}
& \multicolumn{3}{c}{\textbf{Evidence}}
& \multicolumn{2}{c|}{\textbf{Memory}}
& \multirow{2}{*}{Avg.} \\
\cmidrule(lr){2-4} \cmidrule(lr){5-6}
& L1 & L2 & L3 & Scene & Agent & \\
\midrule
\multicolumn{7}{l}{\textit{Spatial evidence ablation}} \\
VLM-only
&  &  &  &  &  & 45.6 \\
+ Level-1 2D evidence
& \checkmark &  &  &  &  & 49.0 \\
+ Level-2 3D evidence
& \checkmark & \checkmark &  &  &  & 49.8 \\
+ Level-3 3D experts
& \checkmark & \checkmark & \checkmark &  &  & 56.7 \\
\midrule
\multicolumn{7}{l}{\textit{Memory ablation}} \\
Spatial only
& \checkmark & \checkmark & \checkmark &  &  & 56.7 \\
+ Scene memory
& \checkmark & \checkmark & \checkmark & \checkmark &  & 58.2 \\
+ Agent memory
& \checkmark & \checkmark & \checkmark &  & \checkmark & 57.6 \\
Full \textsc{S-Agent}
& \checkmark & \checkmark & \checkmark & \checkmark & \checkmark & \textbf{60.0} \\
\bottomrule
\end{tabular}%
}
\end{minipage}
\vspace{-3mm}
\end{table*}

\subsection{Trajectory Distillation from S-Agent}
\label{sec:experiments:training}

We evaluate whether the reasoning trajectories generated by \textsc{S-Agent} can be used to train a smaller open-weight spatial agent. 
Specifically, we fine-tune Qwen3-VL-8B-Instruct on \textsc{S-300K} and obtain \textsc{S-Agent-8B}. 
Table~\ref{tab:trajectory_distillation} compares \textsc{S-Agent-8B} with proprietary VLMs, the original Qwen3-VL-8B-Instruct, and \textsc{S-Agent} using the same Qwen3-VL-8B backbone.
A key observation is that simply equipping the base Qwen3-VL-8B-Instruct with \textsc{S-Agent} does not consistently improve performance. 
The base 8B planner often struggles with tool selection and noisy tool observations, so tool use can bring limited gains or even hurt performance.

In contrast, \textsc{S-Agent-8B} consistently improves over both the base Qwen3-VL-8B-Instruct and \textsc{S-Agent} with the same 8B planner across the three main benchmarks. 
This shows that trajectory distillation teaches not only spatial answers, but also reusable tool-use and evidence-integration patterns for spatial reasoning. 
Notably, \textsc{S-Agent-8B} also achieves competitive performance compared with state-of-the-art proprietary models such as GPT-5.4 and Gemini 3 Pro.

\subsection{Ablation Studies}
\label{sec:experiments:ablation}
We ablate the spatial evidence hierarchy and memory modules of \textsc{S-Agent} on ViewSpatial using GPT-5.4 as the planner. 
As shown in Table~\ref{tab:ablation_config}, adding Level-1 2D evidence improves the VLM-only baseline from 45.6\% to 49.0\%, showing that explicit visual grounding provides useful support for spatial reasoning. 
However, directly adding Level-2 3D evidence provides limited benefit. 
We observe that raw 3D evidence often contains dense numerical information, such as camera poses, depth values, and noisy reconstructed points, which can be difficult for the VLM planner to interpret and may even distract it from the task-relevant spatial cues.

In contrast, enabling Level-3 3D experts substantially improves the score to 56.7\%. 
This suggests that 3D evidence becomes most useful when it is filtered and interpreted by specialized experts, which convert noisy geometric outputs into task-oriented measurements, relative positions, or spatial conclusions. 
The memory modules provide additional gains: scene memory improves the score to 58.2\%, agent memory improves it to 57.6\%, and combining both yields the full \textsc{S-Agent} score of 60.0\%. 
These results show that \textsc{S-Agent} benefits from both structured spatial evidence and persistent memory, with expert-mediated interpretation being crucial for effectively using 3D information.

\subsection{Qualitative Analysis}
\label{sec:experiments:analysis}

\begin{figure}
    \centering
    \includegraphics[width=0.95\linewidth]{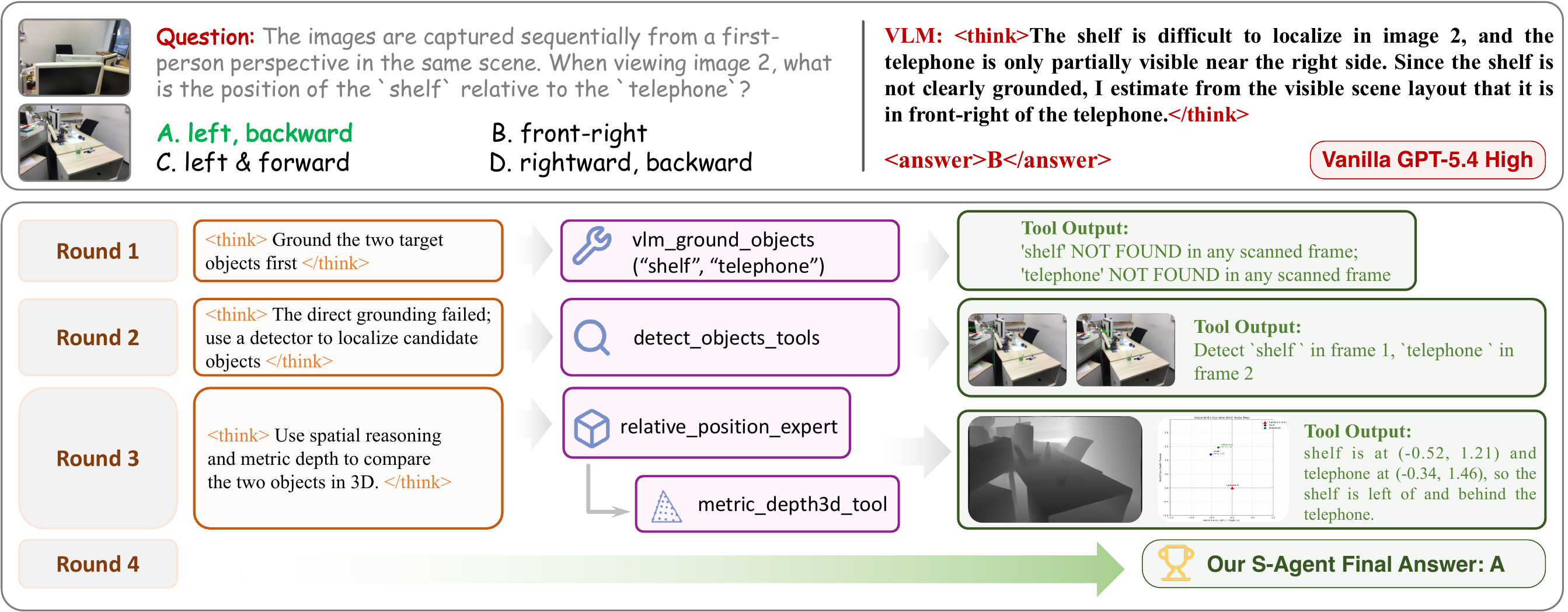}
    \vspace{-2mm}
    \caption{Qualitative example of tool-grounded spatial reasoning. Unlike vanilla VLMs that fail on incomplete cues, our approach accurately infers 3D relations using hierarchical spatial tools and a depth-guided position expert.}
    \label{fig:case_relative}
    \vspace{-2mm}
\end{figure}

We provide qualitative examples to illustrate how \textsc{S-Agent} obtains explicit spatial evidence before answering. Figure~\ref{fig:case_relative} shows a relative-position question in a first-person video. A direct VLM response struggles with this case because the queried objects are partially occluded, and not both clearly visible in the target view. Without grounded evidence, it relies on the apparent 2D layout and incorrectly guesses that the shelf is in the front-right direction.

In contrast, \textsc{S-Agent} follows a tool-grounded trajectory. Although the initial grounding tool fails to locate both queried objects, the agent does not answer immediately. It instead issues targeted detection calls over the video frames, using both the original object names, ``shelf'' and ``telephone'', and a semantically related query, ``desk phone''. These calls recover usable boxes for the shelf and telephone. The relative-position expert then lifts the selected boxes into a metric 3D representation via the depth tool and constructs a bird's-eye-view layout. In this layout, the shelf is estimated at $(-0.52, 1.21)$ and the telephone at $(-0.34, 1.46)$. The recovered geometry shows that the shelf is to the left of and behind the telephone, leading to the correct answer.

\begin{figure}[H]
    \centering
    \includegraphics[width=0.98\linewidth]{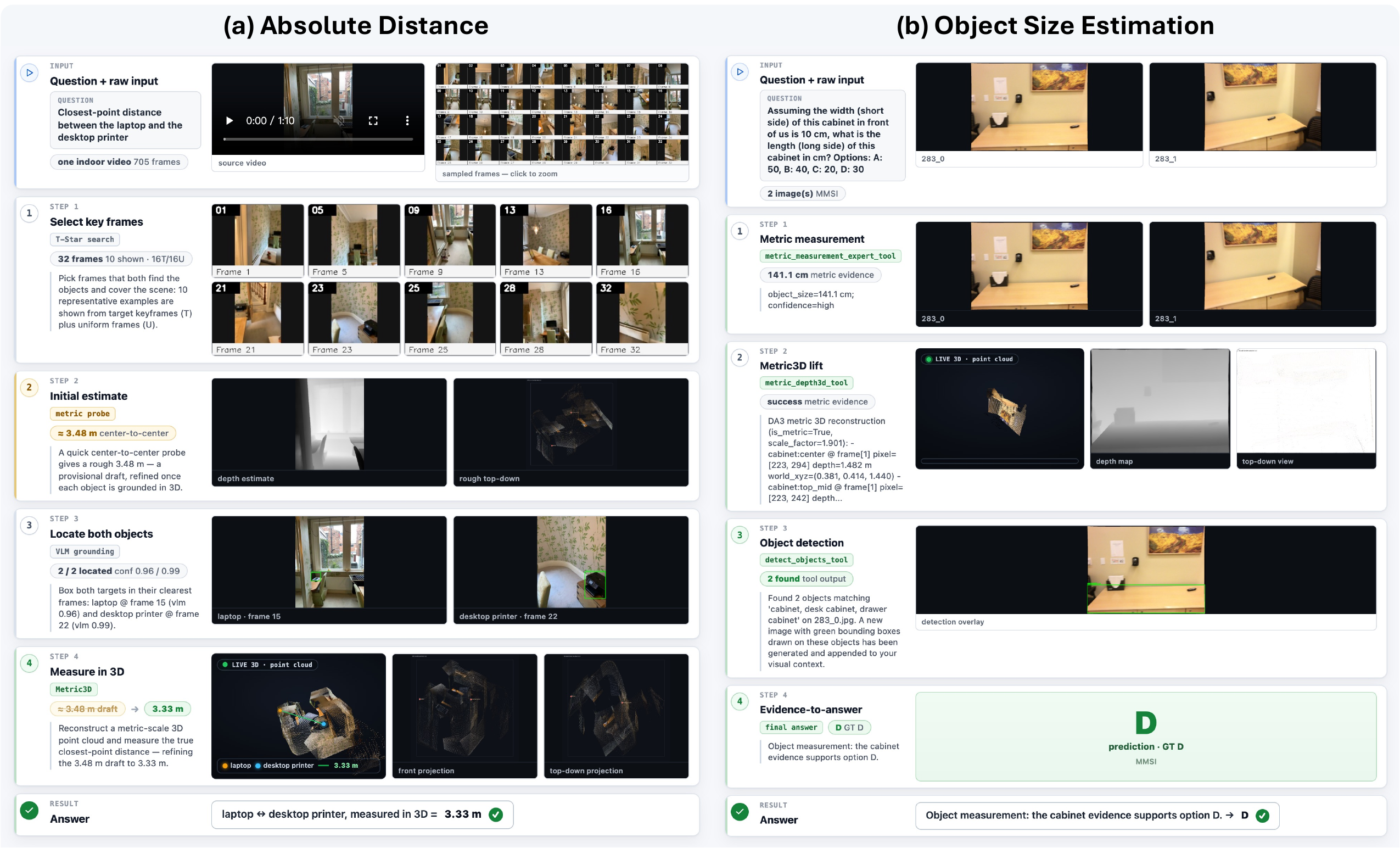}
    \vspace{-2mm}
    \caption{Additional qualitative visualizations of \textsc{S-Agent} across representative spatial reasoning tasks.}
    \label{fig:additional_vis}
    \vspace{-2mm}
\end{figure}

Beyond the detailed case in Figure~\ref{fig:case_relative}, Figure~\ref{fig:additional_vis} provides broader qualitative visualizations across diverse spatial reasoning scenarios, including absolute distance estimation, object size estimation, object counting, multi-step reasoning, relative position reasoning, and route planning. These examples show that \textsc{S-Agent} does not rely on a fixed prompt or a single type of visual cue. Instead, it dynamically invokes different tools and experts according to the task, such as metric measurement for distance and size, key-frame selection and counting tools for object enumeration, 3D lifting for relational reasoning, and route-oriented evidence aggregation for navigation-style questions. Across these cases, \textsc{S-Agent} selects evidence frames, grounds relevant objects, lifts visual observations into metric or top-down spatial evidence, and aggregates the recovered evidence into a final answer.


\section{Conclusion}
\label{sec:conclusion}

We introduce \textsc{S-Agent}, a spatial tool-use agentic framework for spatial reasoning over continuous multi-view images and videos. 
Instead of treating spatial reasoning as a single-shot prediction from isolated visual inputs, \textsc{S-Agent} formulates it as a process of spatio-temporal evidence accumulation. 
It uses a VLM planner to actively acquire hierarchical spatial evidence, from 2D grounding to 3D geometric lifting and expert-level spatial knowledge, while maintaining scene and agent memories for stateful reasoning across views, frames, and tool-use steps. 
Extensive experiments show that \textsc{S-Agent} consistently improves strong VLMs in a training-free zero-shot setting, especially on motion, perspective-aware, and multi-step spatial reasoning tasks. 
Furthermore, trajectories generated by \textsc{S-Agent} can be distilled into \textsc{S-Agent-8B}, enabling an open-weight 8B model to learn more reliable tool-use and spatial evidence integration. 
These results suggest that agentic evidence accumulation is a promising direction for building VLMs with stronger and more grounded spatial intelligence.



\newpage
{
\small
\bibliography{references}
}

\newpage
\appendix

\section*{Appendix}

\section{Related Work}
\label{sec:related}

\paragraph{Spatial Intelligence in VLMs.}
Recent work has sought to improve the spatial intelligence of VLMs by scaling spatial supervision, introducing geometry-aware architectures, or designing spatially focused training objectives. 
Cambrian-S~\citep{cambrian_s} and SenseNova-SI~\citep{sensenova_si} construct large-scale spatial instruction data, while Spatial-MLLM~\citep{spatial_mllm} and VST~\citep{vst} inject explicit spatial modeling or visual spatial tuning into multimodal backbones. 
Other works, such as SpaceR~\citep{spacer}, ViLaSR~\citep{vilasr}, MindCube~\citep{mindcube}, and SpatialLadder~\citep{spatialladder}, further improve spatial reasoning through reinforcement learning, verifiable rewards, or curriculum design. 
These efforts have advanced performance on spatial benchmarks such as BLINK~\citep{blink}, 3DSR~\citep{threedsr}, EmbSpatial~\citep{embspatial}, MMSI-Bench~\citep{mmsi_bench}, and VSI-Bench~\citep{vsi_bench}. 
However, most of them remain \textit{training-driven and single-shot}: the model is expected to encode spatial capability into its parameters and produce an answer in one forward pass, relying on the model's internalized spatial knowledge rather than explicit, scene-specific evidence acquisition at inference time.

\paragraph{Agentic Spatial Reasoning.}
Tool-use agents extend language and vision-language models by interleaving reasoning with calls to external tools, as shown in general agent frameworks such as ReAct~\citep{react} and visual tool-use systems such as ViperGPT~\citep{vipergpt}, Visual ChatGPT~\citep{visual_chatgpt}, and MM-ReAct~\citep{mm_react}. 
More recent work brings this paradigm to spatial reasoning by equipping VLMs with explicit geometric tools or structured computation. 
VADAR~\citep{vadar} synthesizes Python programs over dynamically constructed 3D APIs, SpaceTools~\citep{spacetools} trains VLMs to coordinate vision and robotic tools through reinforcement learning, and GCA~\citep{gca} constrains the reasoning process with formal reference-frame and objective constraints before deterministic geometric computation. 
Concurrent to these efforts, Think3D~\citep{zhang2026think3d} equips VLM agents with 3D reconstruction and camera-manipulation tools, enabling active exploration through ego/global-view switching and novel-view rendering. 
These methods demonstrate the promise of agentic spatial reasoning, but they are still limited in capturing the continuous nature of human spatial understanding, where partial observations are integrated over time, object states are maintained across viewpoints, and spatial judgments are made from an evolving scene representation.

\paragraph{Long-video and Multi-view Understanding.}
Methods commonly handle continuous observations through frame compression or reconstruction-first pipelines. 
Frame-compression methods sample, retrieve, or summarize a limited set of frames before feeding them to long-context VLMs~\citep{videollama,longva,tstar}, improving efficiency but risking the loss of question-relevant spatial evidence. 
Reconstruction-first methods instead build an explicit 3D representation using multi-view geometry or feed-forward reconstruction models~\citep{vggt,da3}, providing stronger geometric grounding but often incurring unnecessary computation when the query only requires sparse or localized evidence. 
However, the selected frames or reconstructed geometry are typically consumed as fixed context, leaving spatial grounding, cross-view association, and metric comparison largely to implicit reasoning or a separate downstream step. 
Thus, they improve access to visual or geometric information, but do not fully close the loop between evidence acquisition, spatial computation, and persistent scene-level reasoning.

\section{Details of Tools and Experts}
\label{app:tools}

\paragraph{Level 1 tools.}
Level 1 contains tools for extracting query-relevant evidence from raw 2D observations. 
The \texttt{detect\_objects\_tool} performs open-vocabulary 2D object detection using GroundingDINO. 
Given an image path and a text prompt, it returns bounding boxes, confidence scores, predicted labels, textual location descriptions, and a visualization with detected boxes. 
This tool converts entities mentioned in the question into localized 2D regions, which serve as the basis for later measurement, counting, and relative-position reasoning.

The \texttt{vlm\_ground\_objects} tool performs multi-frame or multi-image grounding for target entities. 
It uses a two-stage procedure: first, a VLM performs a visibility vote over candidate frames to determine where the target is visible; second, \texttt{detect\_objects\_tool} is applied to the selected best frame to obtain the final bounding box. 
The tool returns the best supporting frame, bounding box, VLM confidence, detector confidence, and visualization for each target. 
This is useful when the target may appear across multiple views or when the referring expression is too complex for direct single-frame detection.

The \texttt{depth\_estimation\_tool} provides lightweight image-level depth cues. 
Given a single image and optional query points, it returns a depth-map visualization and depth estimates at the specified locations. 
We use this tool to support simple depth, occlusion, and front/back reasoning at the image level. 
It should be distinguished from Level 2 geometric lifting, as it provides local image-level depth evidence rather than a full 3D scene representation.

For videos, \textsc{S-Agent} further uses frame or keyframe selection tools, such as \texttt{TStarKeyframeSearchTool}, to identify informative frames before applying the above image-level tools. 
This reduces redundant visual input and allows the agent to focus subsequent grounding and perception on frames that are most relevant to the current question.

\paragraph{Level 2 tools.}
Level 2 contains tools for lifting localized 2D evidence into metric 3D geometry. 
The main tool in our current implementation is \texttt{metric\_depth3d\_tool}, which is built on Depth-Anything-3. 
Given multiple images and query points or boxes, it estimates metric depth, 3D coordinates, camera poses, and depth visualizations. 
This tool provides the shared 3D geometric substrate used by downstream spatial experts, especially the Metric Measurement Expert and Relative Position Expert. 
In our implementation, \texttt{metric\_depth3d\_tool} is the core Level-2 module for stable 2D-to-3D lifting.

\paragraph{Level 3 experts.}
Level 3 consists of five specialized spatial experts: the Metric Measurement Expert, Counting Expert, Visual Orientation Expert, Relative Position Expert, and Object-Centric View Expert. 
Each expert integrates the 2D evidence from Level 1 and, when needed, the lifted 3D evidence from Level 2 to produce structured, scene-specific spatial knowledge for the planner.

\begin{itemize}
\item  \textbf{Metric measurement expert} serves as a geometry-grounded measurement specialist that estimates explicit spatial quantities (e.g., camera-to-object distance, object-to-object distance, and physical object size). 
Given target entities specified by the planner, it first reuses or obtains Level-1 evidence as normalized object boxes, and then queries the Level-2 geometric module to recover metric 3D points inside these regions.
The expert deterministically maps the request to a measurement route, such as \textit{closest-point distance}, \textit{center-to-center distance}, or \textit{longest object dimension}, samples representative points from the grounded boxes, and computes the final value from their recovered 3D coordinates.
It returns a structured observation containing the \texttt{measurement type, numerical value, unit, confidence, and supporting regions}.

\item \textbf{Counting expert} serves as a detection-grounded aggregation specialist that answers object-counting queries, including single-object counts and condition-aware counts over multiple frames.
Given target entities or counting constraints specified by the planner, it first reuses or obtains Level-1 evidence by localizing candidate objects with open-vocabulary detection.
The expert then normalizes the detected boxes across frames, removes duplicated detections with non-maximum suppression, and aggregates the remaining candidates according to the \textit{question-specific counting} target.
For \textit{relational or attribute-conditioned counting}, it further uses the available visual or geometric evidence to filter candidates before computing the final count.
It returns a structured observation containing the \texttt{counted target, numerical count, aggregation mode, confidence, and supporting detections}.

\item \textbf{Visual orientation expert} serves as an appearance-grounded orientation specialist that answers questions about the intrinsic facing direction or pose of an object.
Given the target object and the original question specified by the planner, it collects the relevant Level-1 visual evidence, such as frames where the object is visible and localized object regions when available.
The expert then examines orientation cues including object front/back surfaces, handles, screens, openings, symmetry, and surrounding reference context, and maps the observed pose to the candidate directions or options in the question.
Unlike geometric relation experts that compare object positions in 3D space, this expert focuses on the object's own visual orientation.
It returns a structured observation containing the \texttt{predicted orientation, confidence, and supporting visual evidence}.

\item \textbf{Relative position expert} serves as a 3D relation specialist that answers directional queries between entities, such as left/right, front/back, and cardinal directions.
Given the target and reference entities specified by the planner, it first reuses or obtains Level-1 evidence as grounded object boxes, and then queries the Level-2 geometric module to lift these regions into a shared 3D coordinate system.
The expert deterministically maps the question to a relation route, such as \textit{object-to-object direction, egocentric left/right, viewpoint-conditioned direction, or cardinal-anchor reasoning}.
It then compares the recovered 3D positions under the corresponding reference frame, optionally using camera poses or known direction anchors to calibrate the axes.
It returns a structured observation containing the predicted \texttt{relation or option, confidence, route type, and supporting geometric evidence}.

\item \textbf{Object-centric view expert} serves as a view-aware specialist for questions where the input images are organized around different views of the same target object.
Given the target object, labelled viewpoints, and question context specified by the planner, it reuses Level-1 visual evidence from the corresponding object-centric frames and identifies how surrounding objects appear under the specified viewpoint.
The expert maps the labelled views, such as front, back, left, and right, to the spatial frame required by the question, and then determines the queried relation from this object-centered coordinate system.
It returns a structured observation containing the \texttt{predicted view-conditioned relation, confidence, and supporting frames}.
\end{itemize}

\section{Details of \textsc{S-300K}}
\label{app:training_data}

We construct \textsc{S-300K} from SenseNova-SI-800K~\citep{sensenova_si}, which is fully disjoint from all evaluation benchmarks used in this work. 
The construction pipeline consists of three stages: trajectory generation, trajectory filtering, and trajectory decomposition.

\paragraph{Trajectory generation.}
We follow Section~\ref{sec:method:training} and sample 100K questions from SenseNova-SI-800K and run zero-shot \textsc{S-Agent} with GPT-5.4 as the planner to generate tool-use reasoning trajectories. 
Each trajectory contains the original question, the visual inputs, intermediate planner responses, issued tool calls, returned tool observations, and the final answer produced by the agent.

\paragraph{Trajectory filtering.}
We keep only trajectories whose final answers are valid and correct under the corresponding answer type. 
Specifically, we discard a trajectory if its execution status is marked as failed, if any unrecovered error occurs, or if no final answer is produced. 
For multiple-choice questions, we extract the prediction only from the final \texttt{<answer>...</answer>} field and require the predicted option letter to exactly match the ground-truth option letter. 
For example, if the ground truth is ``B. Northwest'', the trajectory is retained only when the final answer predicts option \texttt{B}. 
For numeric questions, we parse floating-point values from both the prediction and the ground truth, and compute mean relative accuracy (MRA) with a default threshold of 0.6; the trajectory is retained only if $\mathrm{MRA} \geq 0.6$. 
For free-form text questions, we normalize both the prediction and the ground truth by lowercasing, stripping punctuation and extra whitespace, and then require either exact string match or that the ground-truth answer appears as a substring of the predicted phrase. 
In short, our SFT data includes only trajectories whose final answers pass answer-type-specific quality checks. 
We do not use whether a trajectory calls tools as a hard filtering criterion.

\paragraph{Trajectory decomposition.}
After filtering, we decompose each retained trajectory into three complementary supervision formats. 
First, we construct \emph{final-answer trajectories}, where each original question corresponds to one full trajectory ending in the final answer; these samples train the model to imitate complete \textsc{S-Agent} reasoning. 
Second, we construct \emph{turn-level trajectories}, where each VLM planner call is converted into an independent training sample. 
This reduces excessively long contexts, especially for trajectories involving many images or long tool histories, and exposes the model to intermediate planning decisions. 
Third, we construct \emph{expert trajectories}, where individual expert or tool calls are converted into specialized sub-samples, such as calls to the metric measurement expert, counting expert, and relative-position expert. 
A sub-sample is included only when its input is complete, its tool response is available, and the corresponding result can be verified.

\paragraph{Dataset statistics.}
After trajectory generation, filtering, and decomposition, the initial 100K sampled questions yield 292,391 supervised fine-tuning samples. 
We denote the resulting dataset as \textsc{S-300K}. 
Table~\ref{tab:s300k_statistics} summarizes the data statistics. 
Starting from 100,000 raw agent traces, 51,596 trajectories pass the quality filtering stage. 
We keep one final-answer trajectory for each filtered trace, resulting in 51,596 final-answer samples. 
Trajectory decomposition further produces 154,590 turn-level planner samples and 86,205 nontrivial tool/expert samples. 
Together, these three categories form \textsc{S-300K}, containing 292,391 supervised fine-tuning samples. 

\begin{table}[h]
\centering
\small
\begin{tabular}{lc}
\toprule
\textbf{Data Type} & \textbf{Number of Samples} \\
\midrule
Quality-filtered trajectories & 51,596 \\
\midrule
Final-answer trajectories & 51,596 \\
Turn-level trajectories & 154,590 \\
Nontrivial tool/expert trajectories & 86,205 \\
\midrule
Total SFT samples in \textsc{S-300K} & 292,391 \\
\bottomrule
\end{tabular}
\vspace{0.2cm}
\caption{Statistics of \textsc{S-300K}. The main training set consists of final-answer, turn-level, and nontrivial tool/expert trajectories.}
\label{tab:s300k_statistics}
\end{table}

\section{More Experiments}
\label{app:more_experiments}

\paragraph{Results on VSR.}
Table~\ref{tab:vsi_super} shows that \textsc{S-Agent} substantially outperforms existing methods on the VSR subset, achieving particularly large gains in long-video settings. 
For example, under the 240-minute setting, \textsc{S-Agent} surpasses the strongest Cambrian-S-7B-LFP baseline by 37.2 percentage points, which we attribute to the introduction and strong performance of our frame-selection tool. 
On VSC, \textsc{S-Agent} does not outperform Cambrian-S-7B-LFP, but it still performs better than the non-LFP baselines on average.
However, since the reliability of VSI-SUPER as an indicator of genuine spatial perception has been questioned in recent work~\citep{udandarao2025solving}, we avoid over-interpreting this result and include it mainly as a reference.

\begin{table}[t]
    \centering
    \small
    \setlength{\tabcolsep}{8pt}
    \caption{Comparison with existing long-video methods on VSI-SUPER. We report results on VSR and VSC under different video durations.}
    \label{tab:vsi_super}
    \begin{tabular}{l|ccccc|cccc}
        \toprule
        \multirow{2}{*}{\textbf{Eval Setups}} 
        & \multicolumn{5}{c|}{\textbf{VSR (Duration in Mins.)}} 
        & \multicolumn{4}{c}{\textbf{VSC (Duration in Mins.)}} \\
        \cmidrule(lr){2-6} \cmidrule(lr){7-10}
        & \textbf{10} & \textbf{30} & \textbf{60} & \textbf{120} & \textbf{240}
        & \textbf{10} & \textbf{30} & \textbf{60} & \textbf{120} \\
        \midrule
        MovieChat      & 18.3 & 21.7 & 16.7 & 26.7 & 25.6 & 0.0  & 0.0  & 0.0  & 0.0  \\
        Flash-VStream  & 28.3 & 33.3 & 23.3 & 28.3 & 31.7 & 0.0  & 0.0  & 0.0  & 0.0  \\
        Cambrian-S-7B  & 38.3 & 35.0 & 6.0 & 0.0 & 0.0 & 0.0  & 0.0  & 0.0  & 0.0 \\
        Cambrian-S-7B-LFP  & 45.0 & 41.7 & 40.0 & 40.0 & 40.0 & \textbf{40.6} & \textbf{42.0} & \textbf{35.0} & \textbf{34.0} \\
        \midrule
        \textbf{S-Agent (Ours)}  & \textbf{75.0} & \textbf{55.0} & \textbf{63.3} & \textbf{66.1} & \textbf{77.2} & 10.6 & 4.2 & 0.0 & 0.0 \\
        \bottomrule
    \end{tabular}
\end{table}

\section{Additional Qualitative Visualizations}
\label{app:additional_qualitative}

\begin{figure}[p]
    \centering
    \includegraphics[width=\linewidth]{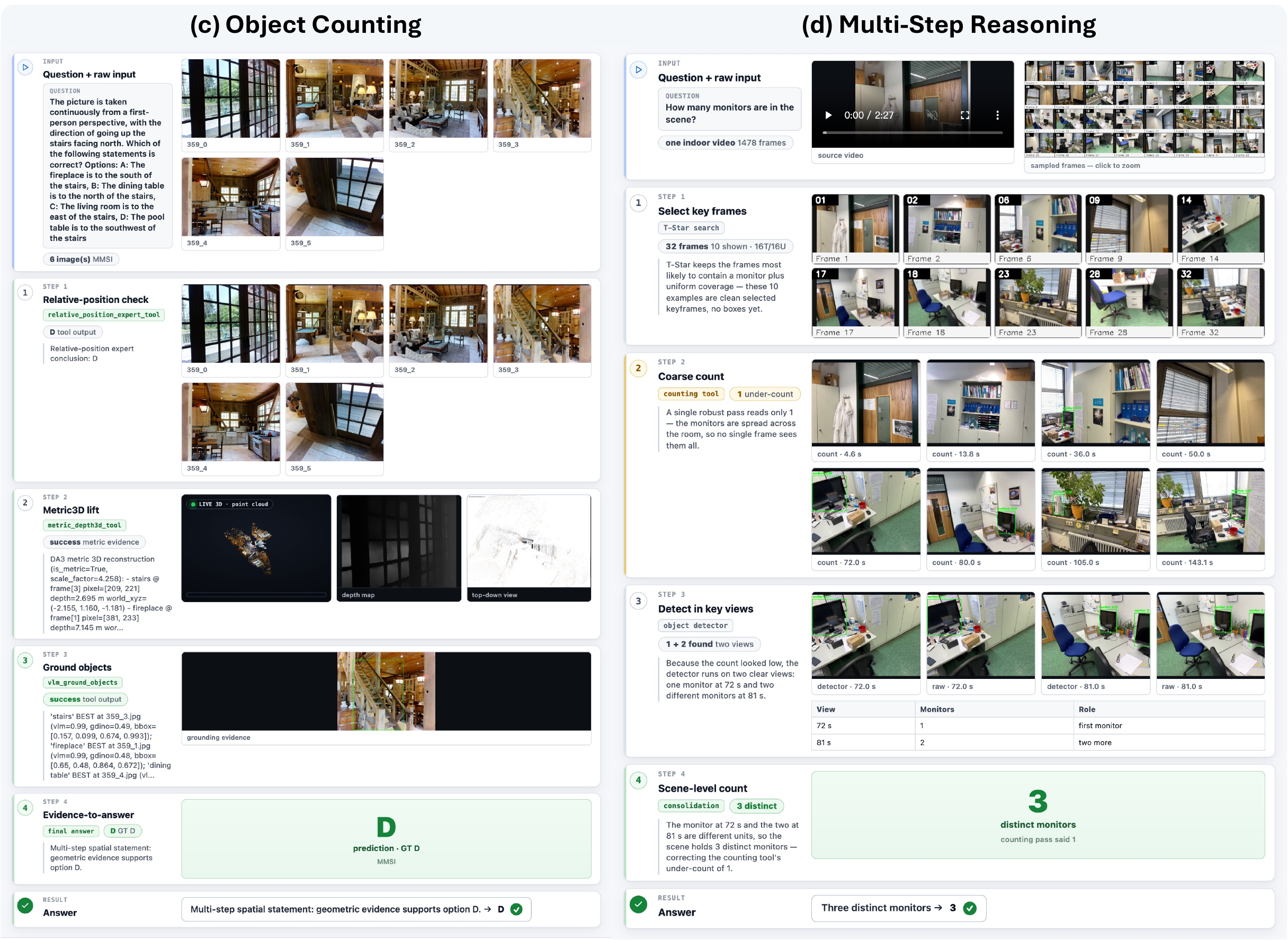}
    \vspace{-2mm}
    \caption{Additional qualitative examples of \textsc{S-Agent} in the appendix.}
    \label{fig:additional_vis_2}
    \vspace{-2mm}
\end{figure}

\begin{figure}[p]
    \centering
    \includegraphics[width=\linewidth]{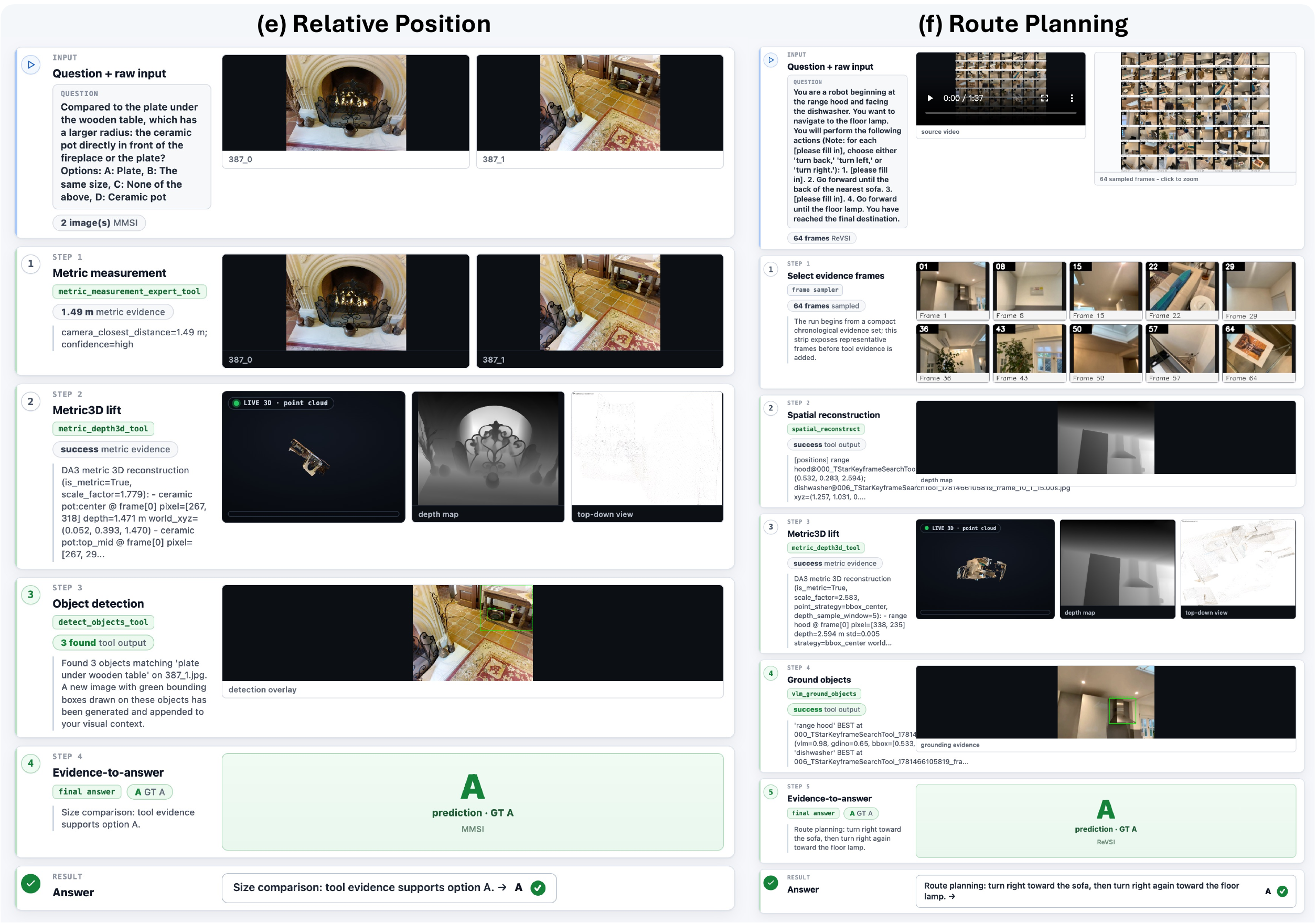}
    \vspace{-2mm}
    \caption{More qualitative examples showing evidence-driven spatial reasoning by \textsc{S-Agent}.}
    \label{fig:additional_vis_3}
    \vspace{-2mm}
\end{figure}

Figures~\ref{fig:additional_vis_2} and~\ref{fig:additional_vis_3} provide additional qualitative examples beyond those in the main paper. These cases further illustrate how \textsc{S-Agent} adapts its tool-use trajectory to different spatial questions, including counting, multi-step reasoning, relative position, and route planning. Across these examples, the agent first selects or grounds task-relevant evidence, then applies metric or spatial experts to convert visual observations into explicit intermediate evidence before producing the final answer.

\end{document}